\documentclass[11pt]{article}
\usepackage[preprint]{acl}
\usepackage{times}
\usepackage{latexsym}
\usepackage[T1]{fontenc}
\usepackage[utf8]{inputenc}
\usepackage{microtype}
\usepackage{amsmath,amssymb}
\usepackage{booktabs}
\usepackage{graphicx}

\newcommand{\pEntries}{5{,}059}
\newcommand{\pIncoming}{41}
\newcommand{\pLinkedNo}{626}
\newcommand{\pHoroyCites}{267}
\newcommand{\pHoroyCitesLinked}{249}
\newcommand{\eLinked}{509}
\newcommand{\eLinkedShare}{59}
\newcommand{\eStrict}{358}

\newcommand{\eLinkedYI}{360}
\newcommand{\eLinkedYII}{149}
\newcommand{\eWordsStrict}{91{,}505}

\newcommand{\aModern}{13{,}122}
\newcommand{\aSilverS}{5{,}446}
\newcommand{\aSilverA}{6{,}156}
\newcommand{\aPlaced}{98.9}
\newcommand{\aCerRaw}{3.40}
\newcommand{\aCerProj}{3.13}
\newcommand{\aStrRaw}{3.86}
\newcommand{\aStrProj}{3.21}
\newcommand{\aWerRaw}{12.5}
\newcommand{\aWerProj}{10.8}
\newcommand{\aAnchorSilver}{5{,}472}
\newcommand{\aBridged}{10}
\newcommand{\lgSet}{legacy BPE 0.1, lexicon 0.05, entities 0.05, gazetteer 0.2}

\newcommand{\lgWerP}{10.0}
\newcommand{\lgWerL}{9.9}
\newcommand{\lgWerC}{9.8}
\newcommand{\lgOovP}{21.9}

\newcommand{\lgOovC}{21.6}
\newcommand{\lgNamP}{27.2}

\newcommand{\lgNamC}{27.4}
\newcommand{\lgUOovP}{18.3}
\newcommand{\lgUOovC}{17.8}

\newcommand{\lgT}{0.2}
\newcommand{\lgM}{0.0}
\newcommand{\lgChanges}{99}

\newcommand{\lgBsD}{$-$0.16}
\newcommand{\lgBsLo}{$-$0.33}
\newcommand{\lgBsHi}{$+$0.03}
\newcommand{\lgBsP}{{=}0.052}
\newcommand{\lgRemoved}{140}
\newcommand{\lgLmWords}{4{,}782{,}222}
\newcommand{\gLam}{0.4}
\newcommand{\gNames}{29.9}
\newcommand{\gNamesG}{29.5}

\newcommand{\gWer}{10.8}

\newcommand{\fSrcWer}{18.7}
\newcommand{\fSrcCer}{9.09}
\newcommand{\fRepWer}{16.0}
\newcommand{\fRepCer}{7.37}
\newcommand{\fOraWer}{12.6}

\newcommand{\fRepFinal}{20.5}
\newcommand{\fOraFinal}{14.6}

\newcommand{\fAdaptLines}{1{,}595}
\newcommand{\fAdaptDev}{176}
\newcommand{\fOraLines}{3{,}646}
\newcommand{\fSrcLines}{6{,}879}

\newcommand{\fBsWerD}{$-$2.8}
\newcommand{\fBsWerLo}{$-$3.6}
\newcommand{\fBsWerHi}{$-$1.8}

\newcommand{\fBsCerD}{$-$1.7}

\newcommand{\tProjGainHyb}{3.2}
\newcommand{\tProjGainCtc}{3.4}
\newcommand{\tSeedGainHyb}{2.0}
\newcommand{\tFinalLo}{24}
\newcommand{\tFinalHi}{35}
\newcommand{\tMIhybWer}{23.3}

\newcommand{\tMIctcWer}{26.9}

\newcommand{\tMIIhybWer}{20.1}

\newcommand{\tMIIctcWer}{23.6}

\newcommand{\tMIIIhybWer}{18.1}

\newcommand{\tMIVshybWer}{14.5}

\newcommand{\bCalPairs}{1{,}475}
\newcommand{\bCalAgree}{7.44}
\newcommand{\bCalTrue}{4.87}
\newcommand{\bBias}{$-$2.6}
\newcommand{\bBiasLo}{$-$3.4}
\newcommand{\bBiasHi}{$-$1.9}
\newcommand{\bCorr}{0.17}
\newcommand{\bCov}{48}
\newcommand{\bCovCal}{51}
\newcommand{\bAgree}{10.37}
\newcommand{\bEst}{7.80}
\newcommand{\bAgreeLo}{10.00}
\newcommand{\bAgreeHi}{10.79}
\newcommand{\dpSlots}{1{,}794}
\newcommand{\dpPlace}{824}

\newcommand{\dpDTFone}{56.5}
\newcommand{\dpDTFzero}{39.9}
\newcommand{\dbDTFone}{58.5}
\newcommand{\dbDTFzero}{33.1}
\newcommand{\dpAllFone}{54.9}
\newcommand{\dpGoldDT}{153}
\newcommand{\sGold}{758}

\newcommand{\sGoldNoText}{312}
\newcommand{\sRegests}{970}
\newcommand{\sRepaired}{109}

\newcommand{\sPP}{60.2}
\newcommand{\sPR}{36.3}
\newcommand{\sDF}{58.5}
\newcommand{\sDR}{55.4}
\newcommand{\sUF}{63.7}
\newcommand{\sUR}{68.7}

\newcommand{\sPRText}{37.7}
\newcommand{\sPRNoText}{34.3}
\newcommand{\daUR}{73.9}
\newcommand{\daUF}{65.6}
\newcommand{\daUFlo}{62.5}
\newcommand{\daUFhi}{68.5}
\newcommand{\daURun}{73.4}
\newcommand{\daDR}{$+$5.1}
\newcommand{\daDRlo}{$+$3.5}
\newcommand{\daDRhi}{$+$6.9}

\newcommand{\daAgree}{89}
\newcommand{\daChecked}{779}
\newcommand{\daDated}{877}
\newcommand{\daRegests}{970}
\newcommand{\daMatched}{357}
\newcommand{\daDPF}{54.9}
\newcommand{\dpAllFzero}{40.6}
\newcommand{\dbAllFzero}{35.0}
\newcommand{\dbAllFone}{58.5}
\newcommand{\daMissParse}{73}
\newcommand{\daMissFolio}{56}
\newcommand{\daMissPlace}{67}
\newcommand{\daMissDate}{2}
\newcommand{\sgProp}{46}
\newcommand{\sgHit}{13}
\newcommand{\sgPrecLo}{17}
\newcommand{\sgPrecHi}{43}
\newcommand{\sgN}{66}
\newcommand{\sgCov}{70}
\newcommand{\sgPrec}{28}
\newcommand{\sgPrecOr}{36}
\newcommand{\sgTr}{607}
\newcommand{\sgTrCov}{52}
\newcommand{\sgBl}{454}
\newcommand{\sgBlCov}{83}
\newcommand{\sgBlAmb}{64}
\newcommand{\auWords}{84{,}660}

\newcommand{\auFlags}{419}
\newcommand{\auAcc}{355}
\newcommand{\auRate}{0.42}

\newcommand{\auSR}{75}
\newcommand{\auSZ}{10}
\newcommand{\auSQ}{15}
\newcommand{\auSP}{88}
\newcommand{\auSPlo}{80}
\newcommand{\auSPhi}{93}
\newcommand{\auVtwo}{322}
\newcommand{\hsBound}{191r}
\newcommand{\hsConf}{88}

\newcommand{\hsSides}{562}
\newcommand{\nmAddrB}{1}
\newcommand{\nmAddrA}{3{,}845}
\newcommand{\nmLetB}{3{,}830}
\newcommand{\nmLetA}{3{,}928}

\newcommand{\nmRep}{129}
\newcommand{\cvNever}{46.7}
\newcommand{\cvRegests}{5{,}059}
\newcommand{\cvNeverOne}{27.1}
\newcommand{\cvNeverEleven}{65.4}

\newcommand{\oNames}{441}
\newcommand{\oBest}{70.1}
\newcommand{\oOracle}{78.2}
\newcommand{\oAbsent}{73}

\newcommand{\kNames}{29.7}
\newcommand{\kNamesG}{29.7}

\newcommand{\oInGaz}{139}

\newcommand{\kInGaz}{113}

\newcommand{\adKappa}{0.73}
\newcommand{\adAgree}{98.1}
\newcommand{\adDisagree}{349}
\newcommand{\gbRuleF}{68.5}

\newcommand{\gbAF}{79.3}

\newcommand{\gbAHF}{66.9}

\newcommand{\gbBF}{92.2}

\newcommand{\gbBHF}{81.7}

\newcommand{\gbHeadRW}{11.0}
\newcommand{\gbHeadRF}{60.0}

\newcommand{\gbHeadNF}{70.7}

\newcommand{\gbHeadEW}{11.1}
\newcommand{\gbHeadEF}{82.1}

\newcommand{\gbHeadELF}{80.2}

\newcommand{\gbHeadLF}{66.0}

\newcommand{\gbNames}{692}
\newcommand{\gbPdEB}{$+$0.4}
\newcommand{\gbPdEBLo}{$-$2.4}
\newcommand{\gbPdEBHi}{$+$3.4}

\newcommand{\gbPdEA}{$+$15.2}
\newcommand{\gbPdEALo}{$+$11.6}
\newcommand{\gbPdEAHi}{$+$18.9}

\newcommand{\gbPdER}{$+$22.1}
\newcommand{\gbPdERLo}{$+$18.0}
\newcommand{\gbPdERHi}{$+$27.4}

\newcommand{\gbChanged}{40}
\newcommand{\gbAgreeErrHT}{11}

\newcommand{\gbNAgName}{50}
\newcommand{\gbPopAgName}{414}

\newcommand{\gbErrAgCapO}{24}
\newcommand{\gbNAgCapO}{50}
\newcommand{\gbPopAgCapO}{370}
\newcommand{\gbErrAgCapOLo}{12}
\newcommand{\gbErrAgCapOHi}{36}
\newcommand{\gbDisChecked}{349}
\newcommand{\lxTypes}{408{,}329}
\newcommand{\lxTokens}{5{,}056{,}092}
\newcommand{\lxEnts}{85{,}688}
\newcommand{\lxOovU}{82.2}
\newcommand{\lxOovUs}{93.5}
\newcommand{\lxNameU}{84.5}
\newcommand{\lxNameUe}{90.1}

\newcommand{\cvPages}{134}
\newcommand{\cvSrc}{17.0}
\newcommand{\cvRep}{15.5}
\newcommand{\cvOra}{12.5}
\newcommand{\cvD}{$-$1.50}
\newcommand{\cvDabs}{1.50}
\newcommand{\cvLo}{$-$1.84}
\newcommand{\cvHi}{$-$1.15}
\newcommand{\cvCerD}{$-$0.89}
\newcommand{\cvRawD}{$+$2.2}

\newcommand{\cvGap}{33}
\newcommand{\sdA}{$-$3.1}
\newcommand{\sdB}{$-$2.8}
\newcommand{\sdC}{$-$2.6}
\newcommand{\sdMean}{$-$2.8}

\newcommand{\rrWer}{17.3}
\newcommand{\rrD}{$-$1.5}
\newcommand{\rrLo}{$-$2.6}
\newcommand{\rrHi}{$-$0.3}

\newcommand{\nrYI}{$+$3.9}
\newcommand{\hrD}{$+$0.7}
\newcommand{\hpD}{$-$2.2}

\newcommand{\lrD}{$-$2.7}

\newcommand{\lrYIwer}{13.5}
\newcommand{\mrYIwer}{11.6}
\newcommand{\mgD}{$-$0.5}
\newcommand{\thD}{$-$0.8}

\newcommand{\cvDpre}{$-$1.46}
\newcommand{\ctWer}{17.1}
\newcommand{\ctWerM}{17.0}
\newcommand{\ctTagCer}{10.8}
\newcommand{\ctModCer}{10.7}
\newcommand{\ctFilmWer}{16.2}
\newcommand{\ctFilmTagCer}{10.6}
\newcommand{\ctFilmModCer}{10.7}
\newcommand{\fwEst}{7.64}
\newcommand{\fwLo}{6.68}
\newcommand{\fwHi}{8.52}
\newcommand{\mixBlk}{85}
\newcommand{\mixCal}{84}
\newcommand{\hAgBlk}{9.8}
\newcommand{\hAgCal}{7.1}
\newcommand{\byD}{$-$0.7}
\newcommand{\byLo}{$-$2.5}
\newcommand{\byHi}{$+$1.1}
\newcommand{\fhRaw}{3.45}
\newcommand{\fhProj}{3.12}

\newcommand{\rsEm}{41}
\newcommand{\rsOcr}{28}
\newcommand{\rsEdge}{22}
\newcommand{\rsVar}{7.8}

\newcommand{\vTwoCer}{3.30}
\newcommand{\vOneCer}{3.32}
\newcommand{\vTwoPairs}{5{,}205}
\newcommand{\cvSegAPages}{108}
\newcommand{\cvSegAD}{$-$1.45}
\newcommand{\cvSegALo}{$-$1.84}
\newcommand{\cvSegAHi}{$-$1.04}
\newcommand{\cvSegAOra}{$-$4.19}
\newcommand{\cvSegBPages}{26}
\newcommand{\cvSegBD}{$-$1.70}
\newcommand{\cvSegBLo}{$-$2.32}
\newcommand{\cvSegBHi}{$-$1.07}
\newcommand{\cvSegBOra}{$-$5.82}
\newcommand{\lpRules}{46}
\newcommand{\lpRewrites}{11}
\newcommand{\lpPairs}{3{,}171}
\newcommand{\lpMin}{15}
\newcommand{\lpHeldHand}{3.56}
\newcommand{\lpHeldLearn}{3.38}
\newcommand{\lpYtwoHand}{3.53}
\newcommand{\lpYtwoLearn}{3.34}

\newcommand{\lpReadDlo}{0.15}
\newcommand{\lpReadDhi}{0.18}
\newcommand{\lpTiAll}{1{,}881}
\newcommand{\lpTiKeep}{1{,}793}
\newcommand{\lfFlagP}{66.7}
\newcommand{\lfFlagR}{30.9}
\newcommand{\lfWrongN}{1{,}121}
\newcommand{\lfShare}{3.7}
\newcommand{\lxWrongIn}{69.1}
\newcommand{\lxRightIn}{98.7}
\newcommand{\rtAndP}{93.0}
\newcommand{\rtAndR}{14.4}
\newcommand{\rtAndS}{1.2}
\newcommand{\rtBR}{58.2}
\newcommand{\rtBS}{9.6}

\newcommand{\pcChanged}{123}
\newcommand{\pcFixed}{18}
\newcommand{\pcBroken}{8}
\newcommand{\bmOraK}{16.1}
\newcommand{\bmAbsK}{54}
\newcommand{\bmCtcOra}{19.7}
\newcommand{\bmOraS}{21.8}
\newcommand{\bsLo}{29.5}
\newcommand{\bsHi}{30.4}
\newcommand{\bsN}{10}

\newcommand{\pxU}{68.5}

\title{Mining the Nineteenth Century for the Thirteenth: Register-Scale Forced Alignment and HTR Bootstrapping via Legacy Regesta and Antiquarian Editions}

\author{Tam\'as Kov\'acs \\
  University of Graz \\
  Department of Digital Humanities \\\And
  Daniel Luger \\
  Institut f\"ur \"Osterreichische Geschichtsforschung \\
  University of Vienna}

\begin{document}
\maketitle

\begin{abstract}
Handwritten text recognition (HTR) for the registers of Pope Honorius~III depends on modern expert transcriptions, which exist only for regnal years 1--2 of Reg.\ Vat.~9. For the remaining registers, nineteenth-century scholarship offers Pressutti's regesta and partial editions. On fol.~1--203, where both layers exist, we measure what this legacy layer can contribute, on transcriptions that we first audit against it, and find a division of labor across three tiers. At the orthographic tier, raw editions are at best neutral for adapting a recognizer to the second regnal year and harmful without replay, whereas projection onto curial spelling with replay of modern source lines turns them into zero-label adaptation targets: pooled over all \cvPages{} year-2 sides, the WER falls by \cvDabs{} points (95\% CI \cvLo{} to \cvHi), closing \cvGap\% of the gap to an oracle; the spelling changes at the year split, through the scribe or the transcriber, whereas the hand changes visibly only at fol.~\hsBound. At the lexical tier, the corpus-wide legacy lexicon contains \lxOovU\% of the out-of-vocabulary words of unedited entries and flags recognition errors with \lfFlagP\% precision, and labels from pre-trained medieval NER models, distilled into the recognizer's NER head, reach $F_1$ \gbHeadEF\% on an expert-validated benchmark, level with the best tagger applied after recognition. At the topological tier, Pressutti's regests, the closing \emph{datum} clauses, and the regests' dates converted to the chancery's Roman form find \daUR\% of entry starts in the recognized line stream, and edition agreement estimates a CER of \bEst\% beyond fol.~203. Antiquarian scholarship thus cannot replace modern transcription, but it can steer, check, and segment HTR across the untranscribed registers.
\end{abstract}

\section{Introduction}

The registers of Honorius~III (Reg.\ Vat.~9--13) hold more than 5,000 letters and have no modern edition. A forced-aligned recognizer trained on the expert transcriptions of the first two regnal years reads held-out pages of the same volume with 12.5\% WER, and with 8.7\% WER after re-alignment, bidirectional decoding with BPE targets, and language-model rescoring \citep{kovacs2026htr}. These transcriptions end at fol.~203 of Reg.\ Vat.~9. For the remaining folios, two older kinds of text exist: the regesta of \citet{pressutti1888,pressutti1895}, which record for nearly every entry its folio, letter number, a Latin summary, and the incipit, and full texts printed in the nineteenth century, above all in Horoy's \emph{Opera omnia} \citep{horoy1879}, but also in the MGH selection of \citet{rodenberg1883}, in Raynaldi's \emph{Annales ecclesiastici} \citep{raynaldi1864}, and in Theiner's regional collections \citep{theiner1859,theiner1860,theiner1861,theiner1863,theiner1864}. These texts were made for reading: they modernize spelling, abridge formulae, and expand initials, and their OCR adds its own noise.

Aligning images with existing editions is an established way to obtain ground truth for registers and charters \citep{stutzmann2017,bluche2017,torresaguilar2023}, and models trained on such data reproduce the editorial conventions of their targets \citep{torresaguilar2023,kovacs2026htr}; earlier work on the papal registers used crowd-sourced or expert line transcriptions \citep{nieddu2021,ammirati2023}. What has not been measured is how far antiquarian editions of a register stand from a modern transcription of the same letters, and what that distance costs a recognizer.

Reg.\ Vat.~9, fol.~1--203, carries both the modern transcriptions and this legacy layer, so the legacy route can be measured against modern ground truth before it is used where no such truth exists. We address three research questions. \textbf{RQ1} (orthographic tier): how accurately can nineteenth-century editions be force-aligned to recognized lines and projected onto the spelling of the register, and how much of the gap to the second regnal year do they close when a recognizer is adapted without modern labels (Sections~3--5)? \textbf{RQ2} (lexical tier): how much of the out-of-vocabulary words and proper names of unseen pages do the regesta and editions cover, and can pre-trained medieval NER models together with this legacy lexicon reduce name errors and supervise a recognizer's NER head without a hand-written rule (Section~7)? \textbf{RQ3} (topological tier): on folios without an edition, how well can Pressutti's regesta segment the recognized line stream into entries, and how far can edition agreement estimate recognition quality beyond fol.~203 (Sections~5--6)? Our answer is a division of labor: modern transcriptions train the recognizer, while the legacy layer adapts it, checks its vocabulary, and structures its output.

\section{Data, Recognizers, and Legacy Sources}

\paragraph{Data and recognizers.} We use the 394 pages of Reg.\ Vat.~9 (fol.~1--130 and 137--203) with the expert transcriptions of 860 entries (203,819 words) by D.~Luger (year~1) and A.~Singh Bottanova (first half of year~2); every tenth page is a test page and every tenth a development page, giving 315 training, 40 development, and 39 test pages and 13,122 modern silver lines obtained by forced alignment (10,525 training, 1,313 development, and 1,284 test lines). The recognizers are those of \citet{kovacs2026htr}: (1) C, an eight-block Conformer-CTC encoder (12.5\,M parameters, $d=256$, a four-block convolutional front-end on 64-pixel lines) whose padding masks pass through every convolution, pooling, and batch-normalization layer, so that decoding does not depend on the inference batch; (2) the hybrid, which adds a four-layer Transformer attention decoder (0.3 CTC $+$ 0.7 cross-entropy, 16.8\,M parameters) and a linear token-level NER head ($+$0.5 NER loss); and (3) the R2L-BPE hybrid, which extends the hybrid with a right-to-left decoder over BPE-512 targets and Kneser--Ney language-model rescoring of its 16-best lists. %
An automatic audit of the transcriptions flags \auFlags{} of the \auWords{} words of lines that also have an edition target where the projected edition and an out-of-fold HTR reading agree with each other against the transcription; \auAcc{} words (\auRate\%), on which the two witnesses agree letter for letter, are corrected, and all experiments use the audited text. The audit draws on the editions themselves, but it leaves the headline adaptation estimate of Section~5 essentially unchanged (\cvDpre{} WER points on the original transcriptions, \cvD{} on the audited ones). The authors judged a random sample of 100 corrections: \auSR{} are right, \auSZ{} are not justified (medieval spellings the edition normalized, word-boundary artifacts, an edition misprint), and \auSQ{} cannot be decided without the manuscript (\auSP\% of the decidable, 95\% CI \auSPlo--\auSPhi\%); rules derived from these cases reduce the audit to \auVtwo{} corrections for future work.
Recognized lines are segmented by Kraken and first read by CATMuS Medieval~1.5; error rates compare the reading layer of recognizer and expert text (u/v, i/j, ae/e, ti/ci harmonized), with words aligned per line.

\begin{table}[b]
\centering\small\setlength{\tabcolsep}{4pt}
\begin{tabular}{lr}
\toprule
Edition (Internet Archive OCR) & entries \\
\midrule
Horoy 2 & 357 \\
MGH Epp. saec. XIII 1 & 35 \\
Horoy 4 & 24 \\
Horoy 1 & 22 \\
Horoy 3 & 16 \\
Theiner Codex dipl. 1 & 15 \\
Raynaldi (Baronius-Theiner) 20 & 13 \\
Theiner Hibernorum & 10 \\
Theiner Poloniae 1 & 9 \\
Raynaldi (Baronius-Theiner) 21 & 7 \\
Theiner Slavorum merid. 1 & 1 \\
\midrule
all (best source per entry) & 509 \\
\bottomrule
\end{tabular}

\caption{Transcribed entries of Reg.\ Vat.~9 (years 1--2) linked to a nineteenth-century full text, by the edition that matched best.}
\label{tab:sources}
\end{table}

\paragraph{Pressutti.} The OCR of both volumes (Internet Archive) is parsed into \pEntries{} entries with number, date and place, register reference (book, letter, folio), incipit, and edition citations; the appendix of incoming letters (\pIncoming{} \emph{Lettere de' Principi}) forms a separate list. Through Pressutti's number, which the expert transcriptions record, \pLinkedNo{} of the 860 transcribed entries are linked to their regest, and Pressutti cites Horoy for \pHoroyCites{} of them.
Before parsing, the OCR is normalized (one pair of quote marks, re-joined hyphens, running heads removed), and the register references are read in both orders of folio and letter: this recovers the address of \nmAddrA{} regests (only \nmAddrB{} before normalization, because the parser needs two quoted units and the OCR renders the quote marks in many forms) and the letter number of \nmLetA{} (\nmLetB{} before), and \nmRep{} OCR digit confusions are repaired where they break the register's order. Of Pressutti's \cvRegests{} regests, \cvNever\% name no printed edition (an upper bound, since Pressutti's citations are incomplete) (book~1: \cvNeverOne\%, book~11: \cvNeverEleven\%), which is the share of letters for which only the regest exists.

\paragraph{Editions.} Horoy's letter numbers do not follow the register numbering of the transcriptions, so we link by text. Every text is reduced to a matching layer (lower case, letters only; j$\to$i, v$\to$u, y$\to$i, ae/oe/\ae/\oe$\to$e, \emph{ti}$+$vowel$\to$\emph{ci}, \emph{ch}$\to$\emph{h} with \emph{h} dropped, so that \emph{mihi} and \emph{michi} coincide). Three 120-character windows of each transcribed entry are searched in the full OCR text of eleven volumes, and an entry is linked when two windows match in order at a normalized distance of at most 0.25, or one at most 0.12. This links \eLinked{} entries (\eLinkedShare\%; years 1 and 2: \eLinkedYI{} and \eLinkedYII; Table~\ref{tab:sources}) and recovers \pHoroyCitesLinked{} of the \pHoroyCites{} entries for which Pressutti cites Horoy. For \eStrict{} entries (\eWordsStrict{} expert words), the edition span lies within distance 0.30 of the expert text; %
most of the remainder are correct letters whose spans are interrupted by running heads and footnotes at page breaks (the OCR reads Horoy's two columns in the right order).

\begin{table}[b]
\centering\small\setlength{\tabcolsep}{3pt}
\resizebox{\columnwidth}{!}{\begin{tabular}{lrrrrrr}
\toprule
 & & & median & \multicolumn{2}{c}{classicizing graphs} & \\
\cmidrule(lr){5-6}
Edition & entries & strict & span dist. & \multicolumn{1}{r}{in edition} & \multicolumn{1}{r}{in expert} & etc. \\
 & & & \% & /1,000 w. & /1,000 w. & /1,000 w. \\
\midrule
Horoy & 419 & 299 & 10.3 & 99 & 49 & 4.7 \\
MGH Epp. saec. XIII 1 & 35 & 29 & 8.1 & 47 & 48 & 4.6 \\
Theiner & 35 & 15 & 49.3 & 51 & 49 & 11.0 \\
Raynaldi & 20 & 15 & 16.2 & 88 & 42 & 0.5 \\
\bottomrule
\end{tabular}
}
\caption{Linked entries by edition family: strict links (span distance $\le$0.30), median span distance on the matching layer, classicizing graphs (ae/oe/\ae/\oe, \emph{ti}$+$vowel, \emph{mihi}/\emph{nihil}, j) per 1,000 words in the edition and in the expert transcription of the same entries, and truncations by \emph{etc.}}
\label{tab:noise}
\end{table}

\section{Alignment to Nineteenth-Century Texts}

\begin{table*}[t]
\centering\small\setlength{\tabcolsep}{4pt}
\begin{tabular}{lrrrrrrrr}
\toprule
 & silver & placed & \multicolumn{2}{c}{CER, reading} & \multicolumn{2}{c}{CER, strict} & \multicolumn{2}{c}{WER} \\
\cmidrule(lr){4-5}\cmidrule(lr){6-7}\cmidrule(lr){8-9}
Alignment & lines & \% & raw & proj. & raw & proj. & raw & proj. \\
\midrule
strict links, baseline aligner & 5{,}446 & 98.9 & 3.40 & 3.13 & 3.86 & 3.21 & 12.5 & 10.8 \\
strict links, entry anchoring & 5{,}472 & 98.8 & 3.60 & 3.32 & 4.05 & 3.40 & 12.7 & 10.9 \\
strict links, anchoring + formula bridging & 5{,}472 & 98.8 & 3.60 & 3.32 & 4.05 & 3.40 & 12.7 & 10.9 \\
all links, baseline aligner & 6{,}156 & 98.9 & 3.61 & 3.33 & 4.07 & 3.41 & 12.9 & 11.1 \\
\bottomrule
\end{tabular}

\caption{Experiment~1: silver lines from nineteenth-century texts on the 394 pages of Reg.\ Vat.~9, fol.~1--203 (\aModern{} modern silver lines on the same pages). Placed: the nineteenth-century target lies within reading-layer distance 0.35 of the modern target of the same line. CER and WER of the nineteenth-century target against the modern target, raw and after curial projection, on the reading layer (which harmonizes ae/e, \emph{ti}/\emph{ci}, u/v) and on a strict layer without harmonization.}
\label{tab:exp1}
\end{table*}

The CATMuS reading of each line is aligned to the edition spans of the entries on its folio and the neighbors. We compare the baseline semi-global aligner of \citet{kovacs2026htr} (edit distance with re-anchoring; accepted at distance $\le$0.35, silver at $\le$0.25 and $\ge$15 characters) with two additions. \emph{Entry anchoring} assigns lines in reading order to non-decreasing entries, or to none (marginalia, rubrics, entries without edition), by a monotone Viterbi pass over the per-entry distances, with a bonus where a line contains the entry's incipit. \emph{Formula bridging} inserts after every \emph{etc.} the majority continuation of its three preceding words in the Innocent~III edition, and each line uses the better of the plain and the bridged text. \emph{Curial projection} maps the target to the spelling of the register: ae/oe$\to$e and j$\to$i always, \emph{ti}$\to$\emph{ci} and \emph{mihi}$\to$\emph{michi} where the reader shows the curial form, and optionally a bare initial of the reader for an expanded name.

The legacy texts give \aSilverS{} silver lines from the strict links and \aSilverA{} from all links, about half of the \aModern{} lines that the modern transcriptions give on the same pages (Table~\ref{tab:exp1}), and \aPlaced\% of them are placed on the same text as the modern target. The remaining differences are editorial rather than positional: \aCerRaw\% CER on the reading layer, \aStrRaw\% on the strict layer, and \aWerRaw\% WER, which curial projection lowers to \aCerProj\%, \aStrProj\%, and \aWerProj\%. Anchoring adds few lines (\aAnchorSilver{} against \aSilverS) at slightly lower fidelity, and bridging applies to \aBridged{} lines only, because the register itself abridges most formulae. Placement is therefore solved by the baseline aligner. Projection removes the spelling of the classicizing Horoy edition (\fhRaw\% $\to$ \fhProj\% CER for its lines) and leaves the MGH, Raynaldi, and Theiner lines unchanged. What remains after projection is mostly not spelling: edition emendations and abridgements account for \rsEm\% of the residual edits, OCR noise for \rsOcr\%, and words cut at line edges by the alignment for \rsEdge\%, whereas curial variants that projection does not model (numerals, \emph{eandem}, doubled consonants) account for \rsVar\%, none of them for more than 3\%; projecting all of them as well would lower the reading-layer CER only from \vOneCer\% to \vTwoCer\% (on the \vTwoPairs{} strict-link line pairs where both projections can be compared). Curial projection is thus close to its ceiling. The rules need not be written by hand. Learned from the \lpPairs{} silver lines of the year-1 training pages alone (character rewrites with one character of left and up to two of right context, read off the aligned word pairs and kept when they make at least \lpMin{} more correct than wrong changes), \lpRules{} rules with \lpRewrites{} distinct rewrites recover j$\to$i, æ/ae$\to$e and \emph{mihi}$\to$\emph{michi}, add three curial variants that the hand rules leave out (\emph{quemdam}$\to$\emph{quendam}, \emph{quum}$\to$\emph{cum}, xs$\to$x) and four OCR confusions of the editions (fl$\to$fi among them), and do not learn ti$\to$ci, because the modern transcriptions keep \emph{ti} before a vowel (\lpTiKeep{} of \lpTiAll{} aligned cases in year~1). On the held-out year-1 pages they lower the strict-layer CER to \lpHeldLearn\% against \lpHeldHand\% for the hand rules, and on year~2, transcribed by the other expert, to \lpYtwoLearn\% against \lpYtwoHand\% (Table~\ref{tab:learnedproj}); gating the less certain rules by the reader, as the hand rules gate ti/ci, changes little. Curial projection is thus a procedure that can be estimated from a few hundred paired lines, not a rule list tied to this register. The experiments below use the hand rules; the learned rules are a later check, and since they also lower the reading-layer CER (by \lpReadDlo--\lpReadDhi{} points), the experiments are, if anything, conservative.

\begin{table}[b]
\centering\small
\resizebox{\columnwidth}{!}{\begin{tabular}{lrrrrrr}
\toprule
 & \multicolumn{2}{c}{all pairs} & \multicolumn{2}{c}{year 1, held-out} & \multicolumn{2}{c}{year 2} \\
\cmidrule(lr){2-3}\cmidrule(lr){4-5}\cmidrule(lr){6-7}
Projection & CER$_s$ & WER & CER$_s$ & WER & CER$_s$ & WER \\
\midrule
raw edition text & 3.86 & 12.5 & 4.19 & 13.0 & 4.27 & 14.1 \\
hand-written rules & 3.21 & 10.8 & 3.56 & 11.3 & 3.53 & 12.2 \\
learned, unconditional & 3.04 & 9.9 & 3.38 & 10.4 & 3.34 & 11.3 \\
learned, reader-gated & 3.03 & 9.9 & 3.36 & 10.4 & 3.33 & 11.3 \\
\bottomrule
\end{tabular}
}
\caption{Curial projection written by hand (Section~3) and learned from the silver lines of the year-1 training pages: strict-layer CER (CER$_s$) and WER of the nineteenth-century target against the modern target of the same line, on all strict-link pairs (the basis of Table~\ref{tab:exp1}), on the held-out (development and test) pages of year~1, and on year~2 (rules learned on year~1 only). Reader-gated: rules with training precision below 0.95 are applied only where they bring the word closer to the recognizer's reading.}
\label{tab:learnedproj}
\end{table}

\section{Training on Antiquarian Lines Only}

\begin{table*}[t]
\centering\small\setlength{\tabcolsep}{4pt}
\begin{tabular}{lrrrrrrr}
\toprule
Training lines & lines & WER & CER & final & OOV & $\alpha$ & $\beta$ \\
\midrule
C, M1: raw 19c lines & 4{,}962 & 26.9 & 10.24 & 35.5 & 36.9 & 19.0 & 23.4 \\
C, M2: projected 19c lines & 4{,}962 & 23.6 & 9.93 & 34.2 & 33.2 & 14.0 & 21.3 \\
C, M3: projected 19c + 1,052 modern & 6{,}014 & 21.7 & 9.34 & 31.4 & 32.2 & 12.9 & 18.1 \\
hybrid, M1: raw 19c lines & 4{,}962 & 23.3 & 9.74 & 29.4 & 33.7 & 15.4 & 17.3 \\
hybrid, M2: projected 19c lines & 4{,}962 & 20.1 & 9.57 & 29.5 & 30.5 & 18.2 & 16.5 \\
hybrid, M3: projected 19c + 1,052 modern & 6{,}014 & 18.1 & 8.72 & 23.6 & 29.0 & 11.9 & 14.1 \\
hybrid, M4s: modern lines, same number as M2 & 4{,}962 & 14.5 & 7.46 & 21.3 & 26.2 & 13.8 & 11.3 \\
C, M4: all modern lines (baseline) & 10{,}525 & 14.7 & 7.20 & 20.7 & 25.8 & 10.6 & 13.9 \\
hybrid, M4: all modern lines (baseline) & 10{,}525 & 12.6 & 6.60 & 19.1 & 23.9 & 9.6 & 10.5 \\
\bottomrule
\end{tabular}

\caption{Experiment~2: recognizers trained from scratch on silver lines of the 315 training pages and evaluated on the 39 test pages. M1--M3: nineteenth-century lines, with development lines from the same source (no modern label in training or selection except the 1,052 modern lines of M3); M4s: as many modern lines as M2; M4: all modern lines (the modern-transcription baseline). Final, OOV: WER on clause-final and out-of-vocabulary words; $\alpha$, $\beta$: cadence errors at clause ends \citep{kovacs2026htr}, non-canonical endings read as canonical ($\alpha$) and canonical endings read as non-canonical ($\beta$).}
\label{tab:exp2}
\end{table*}

C and the hybrid are trained from scratch on the nineteenth-century lines of all training pages: raw (M1), projected (M2), and projected plus 1,052 random modern lines (M3), with selection on nineteenth-century development lines. A size-matched control (M4s) trains the hybrid on as many random modern lines as M2 contains. %
Both runs on M3 stalled with the default seed (the loss did not fall for 60 epochs) and were repeated with the next seed.

Legacy lines alone yield a usable but clearly weaker recognizer (Table~\ref{tab:exp2}): the hybrid reaches \tMIIhybWer\% WER on projected and \tMIhybWer\% on raw lines, C \tMIIctcWer\% and \tMIctcWer\%. %
Projection improves the hybrid by \tProjGainHyb{} and C by \tProjGainCtc{} points, and a small modern seed adds \tSeedGainHyb{} more (\tMIIIhybWer\% for the hybrid). With as many modern lines the hybrid reaches \tMIVshybWer\%, so the editorial divergence of the targets costs more (\tMIIhybWer\% against \tMIVshybWer\%) than halving the modern data. Clause-final words suffer most (\tFinalLo--\tFinalHi\% WER), where the editions abridge closing formulae.

\section{Transfer to the Second Regnal Year} %

\begin{table}[b]
\centering\small\setlength{\tabcolsep}{3pt}
\resizebox{\columnwidth}{!}{\begin{tabular}{lrrrrrr}
\toprule
 & lines & \multicolumn{4}{c}{year 2 (target), 13 pages} & year 1 \\
\cmidrule(lr){3-6}
Model & & WER & CER & final & OOV & WER \\
\midrule
source (year 1, modern lines) & 6{,}879 & 18.7 & 9.09 & 25.7 & 36.2 & 11.4 \\
+ raw 19c lines of year 2 & 1{,}595 & 18.8 & 7.54 & 23.6 & 38.2 & 15.3 \\
+ raw 19c lines + year-1 replay & 3{,}190 & 17.3 & 7.45 & 21.5 & 37.2 & 12.1 \\
+ projected 19c lines of year 2 & 1{,}595 & 16.9 & 7.47 & 26.0 & 37.2 & 13.9 \\
+ projected 19c lines + year-1 replay & 3{,}190 & 16.0 & 7.37 & 20.5 & 34.5 & 11.6 \\
oracle: + modern lines of year 2 & 3{,}646 & 12.6 & 6.25 & 14.6 & 30.1 & 12.9 \\
\bottomrule
\end{tabular}
}
\caption{Experiment~4a: a recognizer trained on year~1 (fol.~1--130) adapted to year~2 (fol.~137--203, the part transcribed by the second expert) without modern year-2 labels. True error rates on the 13 year-2 test pages against the expert transcription, and the WER on the 26 year-1 test pages. Lines: training lines of the step.}
\label{tab:exp4a}
\end{table}

\paragraph{Design.} Year~1 (fol.~1--130) is the source domain and year~2 (fol.~137--203) the target. An R2L-BPE hybrid (without language model) is trained on the \fSrcLines{} modern silver lines of the year-1 training pages. It reads all lines of the year-2 training and development pages, and these readings are aligned to the nineteenth-century texts of the entries on those folios, giving \fAdaptLines{} training and \fAdaptDev{} development lines. No modern year-2 label enters training or model selection. The source model is fine-tuned on these lines (learning rate $3\cdot10^{-4}$, $\le$30 epochs, patience 4), raw or projected, with or without an equal number of replayed year-1 modern lines; as an upper bound, on the \fOraLines{} modern year-2 lines. Year~2 differs from year~1 in spelling from the year split onward, which may stem from the scribe or from the transcriber, and in hand only on fol.~191--203 (see \emph{Beyond fol.~203} below).

\paragraph{Results.} The source model reads year~2 with \fSrcWer\% WER and \fSrcCer\% CER (Table~\ref{tab:exp4a}). Projected legacy lines with replay lower these to \fRepWer\% and \fRepCer\%, a paired bootstrap difference of \fBsWerD{} WER points (95\% CI \fBsWerLo{} to \fBsWerHi) and \fBsCerD{} CER points (paired page bootstrap, 10,000 resamples), against \fOraWer\% with modern year-2 lines; this is the best of three training seeds (\sdA, \sdB, and \sdC{} points; mean \sdMean). Since 13 pages are a small sample, we repeat the experiment as three-fold cross-validation over the 67 year-2 folios, adapting on the legacy lines of the other folds' sides and scoring all sides of the held-out fold. Pooled over \cvPages{} sides, the WER falls from \cvSrc\% to \cvRep\% (\cvD{} points, 95\% CI \cvLo{} to \cvHi; CER \cvCerD), against \cvOra\% for the oracle, so the legacy lines close \cvGap\% of the gap; this is the estimate we rely on. The gain is the same where only the transcriber changes (fol.~137--190, \cvSegAPages{} sides: \cvSegAD{} points, 95\% CI \cvSegALo{} to \cvSegAHi) and where the hand changes as well (fol.~191--203, \cvSegBPages{} sides: \cvSegBD{}, CI \cvSegBLo{} to \cvSegBHi), against oracle gains of \cvSegAOra{} and \cvSegBOra{} points: the legacy lines adapt the model to what the two segments share, the year-2 spelling of scribe or transcriber, and the new hand adds no measurable gain of its own, while the modern labels gain more there. %
Without replay, raw legacy lines raise the pooled WER (\cvRawD{} points in the cross-validation) and the model forgets year~1 (\nrYI{} points on the 13 pages); with replay they help a little (\rrWer\%, \rrD{} points, 95\% CI \rrLo{} to \rrHi), and projection adds more. The edition families show why: raw lines of the classicizing Horoy edition worsen the year-2 WER (\hrD{} points), its projected lines improve it (\hpD), and the raw lines of the MGH and Theiner editions, which keep the register's spelling, help even unprojected (\mgD{} and \thD). Replaying year-1 legacy instead of modern lines gives almost the same year-2 gain here (\lrD) but does not prevent forgetting (year-1 WER \lrYIwer\% against \mrYIwer\%); with the original transcription legacy replay gained less, so whether the modern replay labels add to the year-2 gain remains open.
Tagging each target with its convention instead of projecting it (a token \texttt{<mod>} or \texttt{<horoy>} etc.\ before the decoder targets, CTC on the reading layer) reaches \ctWer\% WER on raw tagged lines (\ctWerM\% with modern replay; \ctFilmWer\% when the tag also conditions the decoder layers by FiLM), against \fRepWer\% for projection with replay. The tag does not steer the decoder in any variant (CER against the raw edition text \ctTagCer\% with the edition tag and \ctModCer\% with \texttt{<mod>}; \ctFilmTagCer\% and \ctFilmModCer\% with FiLM), so it cannot select the convention of the output, which the projection rules fix explicitly.
Clause-final words remain the weak point (\fRepFinal\% WER against \fOraFinal\% for the oracle). Projection and replay together make the legacy layer a zero-label adaptation signal worth about a third of the oracle gap, once some forty edited folios are available.

\paragraph{Beyond fol.~203.} For the untranscribed block fol.~204--287, no true error rate can be computed, so we report \emph{edition agreement}. Two cross-fitted R2L-BPE hybrids, each trained on half of the 394 transcribed sides, read the other half and the block; each read line is searched in all nineteenth-century texts, and the CER of the covered lines (distance $\le$0.30, $\ge$25 characters) against the edition is measured. On the development and test pages of fol.~1--203 it covers \bCovCal\% of the lines, and for \bCalPairs{} covered lines with a modern target agreement CER is \bCalAgree\% and true CER \bCalTrue\%: a bias of \bBias{} points (95\% CI \bBiasLo{} to \bBiasHi), stable in aggregate although the line-level correlation is weak ($r=\bCorr$). On the block, \bCov\% of the lines are covered, with \bAgree\% agreement CER (\bAgreeLo--\bAgreeHi\%), and the calibrated estimate is \bEst\% CER against \bCalTrue\% on the transcribed pages. The edition mix of the covered block lines matches that of the calibration pages (Horoy \mixBlk\% against \mixCal\%), agreement CER is higher on the block within each edition family (Horoy \hAgBlk\% against \hAgCal\%), and calibrating each family separately gives \fwEst\% (95\% CI \fwLo--\fwHi\%); the bias does not differ between the two transcribed years (\byD{} points, 95\% CI \byLo{} to \byHi). %
A change-point analysis of all \hsSides{} sides (the line embeddings of an encoder trained only on Innocent~III's registers, an orthographic fingerprint of the readings, and the corrected agreement) finds no change of hand inside the block: its only visual boundary lies before fol.~\hsBound{} (bootstrap support \hsConf\%), so the block continues the hand of fol.~191--203.
Edition agreement therefore gives an aggregate quality estimate for unseen folios at no annotation cost, and the block is genuinely harder to read than the years used for training, not merely covered by noisier editions.

\section{Pressutti as an Entry Segmenter}

\begin{table}[b]
\centering\small\setlength{\tabcolsep}{3pt}
\resizebox{\columnwidth}{!}{\begin{tabular}{lrrrrrrr}
\toprule
 & & \multicolumn{3}{c}{exact line} & \multicolumn{3}{c}{$\pm$1 line} \\
\cmidrule(lr){3-5}\cmidrule(lr){6-8}
Start lines from & predicted & P & R & F$_1$ & P & R & F$_1$ \\
\midrule
datum clause (no Pressutti) & 677 & 37.1 & 33.1 & 35.0 & 61.9 & 55.4 & 58.5 \\
Pressutti address / incipit & 462 & 47.2 & 28.8 & 35.7 & 60.2 & 36.3 & 45.3 \\
union & 1{,}025 & 37.6 & 50.8 & 43.2 & 59.4 & 68.7 & 63.7 \\
Roman date anchor & 527 & 30.2 & 21.0 & 24.7 & 58.4 & 40.6 & 47.9 \\
Pressutti + date anchor & 912 & 35.2 & 42.3 & 38.4 & 57.8 & 58.7 & 58.2 \\
union + date anchor & 1{,}157 & 35.0 & 53.4 & 42.3 & 58.9 & 73.9 & 65.6 \\
monotone DP (Pressutti + datum) & 592 & 46.3 & 36.1 & 40.6 & 63.5 & 48.3 & 54.9 \\
\quad + date anchor, \emph{ut supra} & 592 & 46.3 & 36.1 & 40.6 & 63.5 & 48.3 & 54.9 \\
\bottomrule
\end{tabular}
}
\caption{Experiment~5: entry start lines on the 394 transcribed sides of Reg.\ Vat.~9 (\sGold{} starts found in the reading), read by the cross-fitted hybrids; gold starts from the (audited) expert transcription, used for evaluation only. Date anchor: a line that reads the Roman date of a regest ends that letter.}
\label{tab:exp5}
\end{table}

Where no edition exists, the recognizer's output is an undivided stream of lines, and Pressutti's folio, address, and incipit are its only external structure. We test Pressutti as an entry segmenter on all 394 sides, read by the cross-fitted hybrids (no page is read by a model that saw it). Gold starts are the lines holding the first 40 characters of each expert entry (\sGold, \sGoldNoText{} without nineteenth-century text). The prediction uses Pressutti only: each of the \sRegests{} regests of books 1--2 is searched on the sides of its folio, $\pm$1 (\sRepaired{} missing folio numbers are imputed from neighboring regests and searched $\pm$2), and a start is placed at the line matching its address, or before the line matching its incipit. A baseline without Pressutti places a start after every line with a \emph{datum} clause.

Pressutti alone finds \sPR\% of the starts within one line, at \sPP\% precision (Table~\ref{tab:exp5}), fewer than the \emph{datum} baseline (\sDR\% recall), because a quarter of the transcribed entries cannot be matched to a parsed regest and short incipits are ambiguous. Yet the two signals are complementary: combining both finds \sUR\% of the starts ($F_1$ \sUF\%, against \sDF\% for the baseline). Pressutti's recall is only slightly lower on entries without an edition (\sPRNoText\% against \sPRText\%), so it carries over to the unedited register.

\paragraph{Monotone alignment.} The register orders the entries of each book by letter number, so the regests can also be aligned to the line stream as a sequence. Gaps in the parsed letter numbers become placeholder slots (\dpPlace{} of \dpSlots); each slot--line pair is scored by a closing formula on the previous line and the address and incipit matches (weights and threshold chosen on the training sides), and the best monotone assignment is decoded by dynamic programming. On all 394 sides, the base of Table~\ref{tab:exp5}, it reaches $F_1$ \dpAllFone\% within one line (exact line \dpAllFzero\%), against \dbAllFone\% (\dbAllFzero\%) for the closing formula alone; on the \dpGoldDT{} starts of the development and test sides only, which played no part in choosing the weights, the two reach \dpDTFone\% (\dpDTFzero\%) and \dbDTFone\% (\dbDTFzero\%). %
 The sequence constraint does not add recall, which locates the bottleneck in matching regests to lines rather than in their order. %
The OCR of the regest strings matters little: replacing them by the expert's own address and incipit raises the union recall only to \pxU\%.

\paragraph{Roman date anchor.} Pressutti dates each regest in modern form; converted to the chancery's Roman form (Kalends, Nones, Ides, \emph{pridie}, the bissextile day, the regnal year from 24 July), the dates of \daDated{} of the \daRegests{} regests of books 1--2 agree with Pressutti's own Roman date in \daAgree\% of the \daChecked{} cases where he prints it, and \daMatched{} are found in the recognized lines. A line that reads the date of a regest closes that letter, so the next line starts an entry. Added to the union, this anchor finds \daUR\% of the starts ($F_1$ \daUF\%, 95\% CI \daUFlo--\daUFhi; recall \daDR{} points, CI \daDRlo{} to \daDRhi), and \daURun\% on entries without an edition (Table~\ref{tab:exp5}). In the monotone DP the training sides give the date no weight, and a soft \emph{ut supra} constraint (an identical date may be written \emph{Dat.\ ut supra}) changes nothing ($F_1$ \daDPF\%). The remaining misses are regests that the parser does not reach (\daMissParse), folios that are missing or off by more than one (\daMissFolio), and regests found but placed on another line (\daMissPlace); a misread date accounts for \daMissDate. The chancery's own dates, not better OCR of the regests, are the lever for regest-driven segmentation.

\section{Legacy Vocabulary and Neural NER}

\begin{table}[b]
\centering\small\setlength{\tabcolsep}{3pt}
\resizebox{\columnwidth}{!}{\begin{tabular}{lrrrrrr}
\toprule
Decoding & WER & CER & OOV & names & sigla & sigla, 31 \\
\midrule
joint 16-best (no gazetteer) & 10.8 & 5.5 & 23.3 & 29.9 & 8.2 & 8.2 \\
+ folio gazetteer ($\lambda=0.4$) & 10.8 & 5.5 & 23.0 & 29.5 & 8.2 & 8.2 \\
\midrule
baseline LMs (char, BPE, word) & 10.0 & 5.3 & 21.9 & 27.2 & 7.3 & 8.0 \\
+ legacy LMs, lexicon, folio gazetteer & 9.9 & 5.3 & 21.6 & 27.7 & 7.3 & 8.2 \\
\quad + OOV post-correction & 9.8 & 5.3 & 21.6 & 27.4 & 7.2 & 8.2 \\
\bottomrule
\end{tabular}
}
\caption{Rescoring of the 16-best lists of the R2L hybrid with NER head trained on re-aligned lines (39 test pages). Names: WER on name words; sigla: WER against the reference with name initials as written, on all test lines; sigla, 31: the same on the 31 lines where the initials convention changes the reference. Top: joint score alone and with the folio gazetteer (with a 5-best list the same model has 10.5\% WER). Bottom: the baseline language models, the leak-free legacy layer, and OOV post-correction.}
\label{tab:exp3a}
\end{table}

\paragraph{Folio gazetteer.} A gazetteer of the capitalized words of Pressutti's regests for each folio and its neighbors, rewarded in the rescoring of the 16-best lists of the NER hybrid ($\lambda=\gLam$ on the development lines), changes the name WER from \gNames\% to \gNamesG\% and leaves the WER at \gWer\% (Table~\ref{tab:exp3a}). Two causes limit it. The attention beam collapses on names: \oBest\% of the \oNames{} name words of the test lines are correct in the best hypothesis and \oOracle\% in at least one of the 16, so for \oAbsent\% of the misread names no hypothesis offers the right form; larger lists help only slowly (oracle name WER \bmOraS\% at 16 and \bmOraK\% at 128 hypotheses, with \bmAbsK\% of the misread names still absent), and the CTC prefix beam levels off at \bmCtcOra\%. The regests also name only the principal persons (\kInGaz{} of the name words are in the folio gazetteer, \oInGaz{} with all capitalized regest words and their case forms). Contextual biasing inside CTC prefix beam search, where the beam does not collapse, therefore does not help either: the name WER is \kNames\% without and \kNamesG\% with the folio gazetteer, and with the leak-free entities of the whole legacy layer it stays between \bsLo\% and \bsHi\% in all \bsN{} settings of bias weight and beam width. Regest summaries are too sparse to steer decoding; the useful lexical evidence lies in the full texts.

\begin{table}[b]
\centering\small\setlength{\tabcolsep}{4pt}
\begin{tabular}{lrrrr}
\toprule
Test words & n & exact & stem & entity \\
\midrule
training-OOV words, all & 1{,}432 & 86.2 & 94.1 & -- \\
\quad in entries without edition & 618 & 82.2 & 93.5 & -- \\
\quad year 2 & 469 & 88.3 & 96.2 & -- \\
names (gold), all & 692 & 87.0 & 93.9 & 92.1 \\
\quad in entries without edition & 303 & 84.5 & 93.4 & 90.1 \\
\quad year 2 & 207 & 92.8 & 98.1 & 98.1 \\
\bottomrule
\end{tabular}

\caption{Coverage of test words by the legacy lexicon (all words of the \pEntries{} regests and the eleven edition volumes after curial projection; \lxTokens{} words, \lxTypes{} types; entity: stems of the words tagged as persons or places by the RoBERTa charter tagger or the Span-NER model in these texts). Training-OOV: words of the 39 test pages that are not in the training vocabulary; names: name words of the gold benchmark of Table~\ref{tab:exp7a}. %
 Editions may print the entries in which a word occurs; the rows without edition are the relevant ones for unedited folios.}
\label{tab:exp7b}
\end{table}

\paragraph{Corpus-wide legacy lexicon.} The full legacy layer of all regests and all eleven volumes (\lxTokens{} words, \lxTypes{} types after projection, OCR forms included) covers far more. Of the training-OOV words in test entries that no edition prints, \lxOovU\% occur in it in exactly the same form (\lxOovUs\% by stem), and of the names \lxNameU\% (\lxNameUe\% among the stems of the \lxEnts{} entity types that either tagger finds in it; Table~\ref{tab:exp7b}). It is too permissive to choose between hypotheses (\lxWrongIn\% of the wrong words of C with language-model fusion occur in it, against \lxRightIn\% of its correct ones). Conversely, absence from the legacy lexicon provides a strong diagnostic signal: with the leak-free lexicon of the next paragraph, a word outside it is wrong with \lfFlagP\% precision, and this flag catches \lfFlagR\% of the \lfWrongN{} wrong words while marking \lfShare\% of all words. Combined with the calibrated word posterior of the hybrid (error probability $\ge$0.3), it sends \rtAndS\% of the words to review at \rtAndP\% precision (\rtAndR\% recall); the posterior alone ($\ge$0.2) catches \rtBR\% of the errors when \rtBS\% of the words are routed.

\paragraph{Legacy lexicon in decoding.} For decoding we rebuild the legacy layer without the edition spans of all entries on development and test pages (\lgRemoved{} spans), so that no edition of a held-out letter enters the models; \lgLmWords{} words remain for a BPE 4-gram and a word trigram model. (a) In the 16-best rescoring, these models, the shares of words in the legacy lexicon and of capitalized words among its entities, and the folio gazetteer join the baseline language models (weights on the development rows: \lgSet). (b) A word outside the training vocabulary whose calibrated error probability (from its posterior in the 16-best list) is $\ge$\lgT{} is replaced by a legacy word or entity within edit distance 2 if the hybrid's CTC head scores the changed line at most \lgM{} nats lower. On the test pages (Table~\ref{tab:exp3a}), the WER falls from \lgWerP\% to \lgWerL\% with (a) and to \lgWerC\% with (b), and the OOV WER from \lgOovP\% to \lgOovC\%, while the name WER moves from \lgNamP\% to \lgNamC\% (\lgChanges{} words changed); on the lines of unedited entries, the OOV WER falls from \lgUOovP\% to \lgUOovC\%. Post-correction acts on few words (of \pcChanged{} changed words, \pcFixed{} become correct and \pcBroken{} wrong). The total change of \lgBsD{} WER points (95\% CI \lgBsLo{} to \lgBsHi; page bootstrap; $p\lgBsP$) is small and, on the audited system, not distinguishable from zero; the name WER does not improve, which positions the legacy lexicon as a signal for review rather than for decoding.

\begin{table}[b]
\centering\small\setlength{\tabcolsep}{3pt}
\resizebox{\columnwidth}{!}{\begin{tabular}{llrrrrr}
\toprule
Method & applied to & WER & P & R & F$_1$ & 95\% CI \\
\midrule
(a) heuristic name rule & reference & -- & 88.0 & 56.1 & 68.5 & [63.0, 72.5] \\
(b) RoBERTa charter tagger & reference & -- & 88.0 & 72.1 & 79.3 & [75.7, 82.9] \\
\quad & recognized text & -- & 78.8 & 58.1 & 66.9 & [61.5, 72.1] \\
(c) bi-encoder Span-NER & reference & -- & 95.7 & 89.0 & 92.2 & [90.2, 94.4] \\
\quad & recognized text & -- & 87.7 & 76.4 & 81.7 & [77.5, 85.6] \\
(d) NER head, rule labels & recognized text & 11.0 & 79.7 & 48.1 & 60.0 & [55.4, 64.6] \\
(e) NER head, RoBERTa labels & recognized text & 11.1 & 83.5 & 61.3 & 70.7 & [66.6, 74.9] \\
(f) NER head, ensemble labels & recognized text & 11.1 & 89.7 & 75.7 & 82.1 & [77.5, 86.3] \\
\bottomrule
\end{tabular}
}
\caption{Name words (persons and places) on the 39 test pages against the gold benchmark (\gbNames{} name words, adjudicated and checked by an expert). Rows (b) and (c) on the reference contribute to the benchmark and are not independent of it. WER: recognition WER of the model; heads are evaluated on their own recognized text, mapped to the reference words. CI: page bootstrap of $F_1$ (10,000 resamples).}
\label{tab:exp7a}
\end{table}

\paragraph{Neural NER.} The NER head of \citet{kovacs2026htr} was trained on labels from a heuristic rule (capitalized, not a Collatinus form, not a title). We replace the rule with two pre-trained models for medieval charters: the RoBERTa tagger of \citet{aguilar2022multilingual} (persons, places) and the bi-encoder Span-NER of \citet{kovacs2026twentys} (19 labels, of which we use persons and places). On the test references the two agree on \adAgree\% of the words (Cohen's $\kappa=\adKappa$ for name versus non-name). Their \adDisagree{} disagreements are first decided by explicit philological and documentary rules (lower case in the expert text, \emph{-ensis} adjectives, offices and calendar terms, saints and feasts, ethnonyms, the folio's Pressutti regests, person or place context, abbreviations, Collatinus forms), and an expert then checked every disagreement (\gbDisChecked{} words) and a random sample of \gbNAgName{} of the \gbPopAgName{} words both models tag and \gbNAgCapO{} of the \gbPopAgCapO{} capitalized words both leave untagged, changing \gbChanged{} labels. Agreement is not truth: none of the sampled agreed names was wrong, but \gbErrAgCapO\% of the agreed capitalized non-names were names (95\% CI \gbErrAgCapOLo--\gbErrAgCapOHi\%), \gbAgreeErrHT\% over both agreement strata. The validated labels form the gold benchmark (\gbNames{} name words). Against it (Table~\ref{tab:exp7a}) the heuristic rule reaches $F_1$ \gbRuleF\%, and both models lose recall when applied post hoc to recognized text (RoBERTa \gbAF\% on the reference and \gbAHF\% on recognized text, Span-NER \gbBF\% and \gbBHF\%). Distilling the ensemble labels into the recognizer's own NER head raises its $F_1$ from \gbHeadRF\% with rule labels to \gbHeadEF\% (\gbPdER{} points, 95\% CI \gbPdERLo{} to \gbPdERHi; paired page bootstrap) at a WER of \gbHeadEW\% against \gbHeadRW\%; RoBERTa labels alone give \gbHeadNF\%. The distilled head is ahead of the RoBERTa tagger run post hoc on the same recognized text (\gbPdEA{} points, CI \gbPdEALo{} to \gbPdEAHi) and level with the Span-NER model run post hoc (\gbPdEB{}, CI \gbPdEBLo{} to \gbPdEBHi), while adding only a linear layer to decoding. Labels from the legacy layer (Pressutti names and legacy entities in the training lines) do not improve on the ensemble labels (\gbHeadELF\% together) and alone reach \gbHeadLF\% (rule labels: \gbHeadRF\%). Pre-trained medieval NER thus supplies entity supervision that a hand-written rule cannot, and the recognizer's head absorbs it at no cost in reading accuracy.

\paragraph{Sigla.} The register often names persons by an initial (\emph{B.\ electo}). A standoff layer proposes the full name without altering the recognized text: candidates come from the parenthesized expansions and names of the entry's regest and from its linked edition, ranked by initial, person context (a following title or office), and the words shared with the line around the initial. Where the register writes a person's name in full and the recognizer reads it, masking the name to its initial and resolving it gives a candidate for \sgCov\% of the \sgN{} names, and the first candidate is the written form for \sgHit{} of these \sgProp{} (\sgPrec\%, 95\% Wilson CI \sgPrecLo--\sgPrecHi\%; \sgPrecOr\% when the entry is known). On the transcribed sides, \sgTrCov\% of the \sgTr{} sigla in the recognized lines receive a candidate, and on fol.~204--287 \sgBlCov\% of \sgBl, with several candidates for \sgBlAmb\%. The layer is therefore a candidate list for expert review, not an automatic resolution.

\section{Discussion}

Nineteenth-century scholarship and HTR divide the work across three tiers. At the orthographic tier (RQ1), forced alignment places legacy text correctly, and curial projection removes the editorial spelling of the classicizing editions: raw editions are at best neutral for an adapted recognizer and harmful without replay, whereas projection and replay of modern source lines turn the same editions into zero-label adaptation targets (\cvD{} WER points pooled over all year-2 sides, \cvGap\% of the oracle gap, given some forty edited folios). The year-2 experiment measures a change of spelling at the year split, attributable to the scribe or the transcriber, and a change of hand only on fol.~191--203 (the visual boundary lies at fol.~\hsBound); the gain for a different scribe remains to be confirmed on the later registers. Legacy text cannot replace modern transcription as primary training data (\tMIIhybWer\% against \tMIVshybWer\% WER with as many modern lines), but it can audit it: where the projected edition and an independent reading agree against the transcription, \auSP\% of the decidable corrections in a judged sample hold. At the lexical tier (RQ2), the corpus-wide legacy lexicon covers \lxOovU\% of the OOV words of unedited entries and identifies recognition errors with \lfFlagP\% precision, enough to send \rtAndS\% of the words to review at \rtAndP\% precision; pre-trained medieval NER models, distilled into the recognizer's head, match the best post-hoc tagger on an expert-validated benchmark (\gbHeadEF\% against \gbBHF\% $F_1$), and the regests propose expansions for initials as a review layer. At the topological tier (RQ3), where full texts are absent, Pressutti's regests, the closing \emph{datum} clauses, and the chancery's Roman dates segment the recognized line stream into entries (\daUR\% of starts; the misses lie in regest parsing and folio reconciliation, not in the OCR of the regests), and edition agreement calibrates aggregate recognition error beyond the transcribed folios (\bEst\% estimated CER). For the later registers this suggests a workflow: train on modern transcriptions, adapt to each new register with projected legacy lines, route flagged words to expert review, and cut the output into entries with the regesta and their dates.

\section*{Limitations}
All measurements come from one register volume. The year-2 test set has 13 pages, and the transfer gain varies by about a point across training seeds; the cross-validated estimate over all \cvPages{} year-2 sides is half the 13-page gain, and its folds adapt on only two-thirds of the legacy lines. The nineteenth-century texts cover about half of the transcribed entries, and their coverage beyond fol.~203 is not known in advance. Linking and alignment depend on the OCR of the Internet Archive scans, which we did not correct. %
The NER gold benchmark starts from two pre-trained models: the expert checked every disagreement but only samples of the agreements, so the unchecked agreed words keep an estimated error of \gbAgreeErrHT\%, concentrated in capitalized words that both models leave untagged; its conventions (saints, feasts, and ethnonyms are not entities) are choices that another annotator might set differently, and the legacy lexicon includes the OCR errors of its sources.
The year-1/year-2 split confounds hand and transcriber. The change-point analysis places the visual change of hand before fol.~\hsBound, inside year~2, whereas the spelling of the readings changes at the year split; the transfer experiment therefore measures a change of spelling, attributable to the scribe or to the transcriber, and a change of hand only for fol.~191--203. Split by segment, the cross-validated gain of the legacy lines is the same on both (Section~5), so it does not depend on the change of hand. The encoder embeddings are trained for recognition, not writer identification, so two scribes of the same chancery school may be visually indistinguishable to them. These boundaries are statistical and await a palaeographic check.

\section*{Author Contributions}
\textbf{Tam\'as Kov\'acs}: Conceptualization, Methodology, Software, Validation, Formal analysis, Investigation, Data curation, Visualization, Writing -- original draft, Writing -- review \& editing. \\
\textbf{Daniel Luger}: Resources, Data curation, Writing -- review \& editing.

\section*{Reproducibility}
Code, parsed Pressutti indices, orthographic projection scripts, and the metrics behind every table are available from the authors; the manuscript images are subject to the rights of the Vatican Apostolic Archive.

{\small\setlength{\bibsep}{2pt}
\bibliography{references}}

\end{document}